\documentclass[letterpaper, 10 pt, conference]{ieeeconf}  
\IEEEoverridecommandlockouts                              
\usepackage{graphics} 
\usepackage{epsfig} 
\usepackage{amsmath} 
\usepackage{amssymb}  

\usepackage{array}
\usepackage{hyperref}

\usepackage{booktabs}

\begin{document}

\title{\LARGE \bf
Revisiting Distribution Matching for State-to-Vision Policy Distillation in Laparoscopic Robotic Manipulation
}


\author{Mingyu Sheng$^{1}$,
        Jianan Fan$^{1}$,
        Dongnan Liu$^{1}$,
        Guoyan Zheng$^{2}$,
        Ron Kikinis$^{3}$,
        Weidong Cai$^{1}$
\thanks{$^{1}$Mingyu Sheng, Jianan Fan, Dongnan Liu, and Weidong Cai are with the School of Computer Science, the University of Sydney, Sydney, NSW, Australia, {\tt\small \{mshe0136, jfan6480\}@uni.sydney.edu.au, \{dongnan.liu, tom.cai\}@sydney.edu.au}.}
\thanks{$^{2}$Guoyan Zheng is with the Institude of Medical Robotics, School of Biomedical Engineering, Shanghai Jiao Tong University, Shanghai, China, {\tt\small guoyan.zheng@sjtu.edu.cn}.}
\thanks{$^{3}$Ron Kikinis is with the Department of Radiology, Brigham and Women's Hospital, and Harvard Medical School, Boston, MA, United States,  {\tt\small kikinis@bwh.harvard.edu}.}
\thanks{Corresponding authors: Jianan Fan and Weidong Cai.}}

\markboth{IEEE ROBOTICS AND AUTOMATION LETTERS,~Vol.~11, No.~6, June~2026}%
{Shell \MakeLowercase{\textit{et al.}}: A Sample Article Using IEEEtran.cls for IEEE Journals}


\maketitle
\thispagestyle{empty}
\pagestyle{empty}

\begin{abstract}


End-to-end control policies trained via Reinforcement Learning (RL) have emerged as an alternative to conventional surgical robotic control approaches, with recent studies exploring state-based policies trained on privileged state observations. 
However, such approaches rely on precise physical states that are difficult to obtain reliably in surgical environments. 
Vision-based RL provides an alternative by learning manipulation policies directly from RGB observations, avoiding explicit state estimation and manually engineered perception pipelines while enabling policies to acquire task-relevant visual representations. 
Nevertheless, obtaining vision-based policies via pure RL methods for surgical manipulation remains challenging. Thus, imitation learning (IL) becomes another solution, yet it requires a large number of expert demonstrations, which are difficult to obtain in minimally invasive surgery.
In this work, we propose a framework for state-to-vision policy distillation in laparoscopic robotic manipulation. 
Specifically, a state-based policy with privileged observations is employed as the teacher, and a vision-based policy learns from RGB observations and robot proprioceptive information, which requires less expert demonstrations than IL and introduces superior performance than RL methods.
We conduct extensive experiments on various surgical manipulation tasks in the simulated environment. 
Experimental results show that, in the context of laparoscopic robotic manipulation, training vision-based policies with general-purpose RL methods is challenging, whereas our proposed state-to-vision policy distillation leads to more consistent and stable performance across tasks.


\end{abstract} 


\section{Introduction}
    Dexterous manipulation skills are critical components of surgical robotics tasks such as needle insertion, knot tying, and tissue retraction \cite{Fiorini_2022_Concepts_and}. 
    In recent years, significant progress has been made in applying deep reinforcement learning (RL) to accomplish such complex autonomous or semi-autonomous subtasks \cite{Bendikas_2023_Learning_needle, Haiderbhai_2024_Sim2real_rope}. 
    This progress has been largely enabled by the development of high-fidelity surgical robot simulators \cite{Tagliabue_2020_Soft_tissue, Scheikl_2023_Lapgym_an, Schmidgall_2024_Surgical_gym, Varier_2022_Ambf_rl, Xu_2021_Surrol_An, Yu_2024_Orbit_surgical}. 
    However, a key limitation of RL-based surgical robotics lies in its reliance on privileged state information, which typically assumes accurate and noise-free access to the physical states of surgical objects, with limited integration of visual perception \cite{Long_2025_Surgical_embodied}. 
    This issue is particularly pronounced in fine manipulation tasks such as needle handling, where even minor deviations in object pose can lead to task failure \cite{Bendikas_2023_Learning_needle}.

    Recently, benchmarks have been proposed to provide a diverse set of learning-based tasks that capture varying levels of dexterity and precision required in the surgical context \cite{Xu_2021_Surrol_An}. 
    Benchmarks provide standardized platforms for evaluating autonomous policies. Nevertheless, surgical manipulation tasks still remain highly challenging due to their stringent requirements for precision, robustness, and safety in constrained operating environments.


    To reduce the reliance on privileged state information, vision-based policy learning provides an alternative by directly mapping visual observations to control actions \cite{Laskin_2020_Curl_Contrastive, Kostrikov_2020_Image_augmentation, Yarats_2021_Mastering_visual}. 
    Such policies can avoid state estimation and manually engineered perception pipelines by leveraging visual information to learn task-relevant representations. 
    However, learning effective control policies directly from visual observations via RL remains a major hurdle in surgical robotics.
    In the context of surgical manipulation tasks, the difficulty is further amplified by the severe visual sparsity of task-relevant objects. 
    For instance, the needle and instruments of surgical robots often occupy only a few pixels in high-resolution RGB observations, making reliable perception and localization hard to achieve. 
    Furthermore, the instability of RL exacerbates the training difficulty in such sparse and precision-critical settings \cite{Henderson_2018_Deep_reinforcement}. 
    Imitation learning (IL) defines a stable training objective, which transforms RL into supervised learning given a large number of expert demonstrations as data labels \cite{Fu_2024_Mobile_aloha, Chi_2025_Diffusion_policy}.
    However, compared with general robotic manipulation tasks, such as opening bottles or folding clothes, expert demonstrations for laparoscopic manipulation tasks are much more expensive and scarce.


    To facilitate vision-based policy learning under these limitations, policy distillation provides an effective approach for transferring knowledge from a state-based policy to an RGB-based policy \cite{Liu_2022_Distilling_Motion, Rusu_2015_Policy_distillation, Czarnecki_2019_Distilling_Policy}. 
    The state-based policy has access to privileged robot and object state information and serves as the teacher, and the RGB-based policy then learns to reproduce the teacher's behavior from visual observations and robot proprioceptive states.
    Such knowledge transfer provides guidance for learning visuomotor policies from high-dimensional visual inputs, enabling the vision-based policy to acquire effective manipulation behaviors without relying on privileged state information during execution.
    However, most studies on policy distillation consider discrete action settings, such as those in Atari games, and thus the potential of policy distillation for continuous-control surgical robotic manipulation remains unexplored.


    In this work, we propose a policy distillation framework for continuous-control laparoscopic robotic manipulation. 
    Specifically, we proposed a distribution-matching-based policy learning approach that overcomes the limitation of previous methods by Gaussian alignment. We demonstrate the effectiveness of the approach through both theoretical analysis and empirical validation. 
    We employ SurRoL \cite{Xu_2021_Surrol_An} as the simulation environment, where a Patient Side Manipulator (PSM) executes the manipulation task while an Endoscopic Camera Manipulator (ECM) provides RGB observations. 
    During training, the state-based policy has access to all privileged state information, including the coordinates and orientations of all objects, and serves as the teacher for the vision-based policy. 
    In contrast, the vision-based policy only receives RGB observations and robot proprioceptive states, and is trained to distill the behavior of the state-based teacher without accessing privileged object states during execution.
    Through theoretical analysis and empirical evaluation, we provide insights into various distribution matching objectives for policy distillation, which may serve as a useful reference for future research on vision-based policy learning in surgical robotics manipulation.
    The main contributions of this work are summarized as follows:
    \begin{itemize}
    \item We formulate state-to-vision policy distillation as a stochastic distribution matching problem, where a vision-based student policy learns to match both the action and the uncertainty of a privileged state-based teacher policy, reducing the demand for expert demonstrations and demonstrating stronger stability.

    \item We provide a unified first- and second-order gradient analysis of commonly used distribution matching objectives, revealing their distinct gradient scaling, parameter coupling, and curvature properties for Gaussian policies.


    \item We evaluate the proposed method across various surgical robotic manipulation tasks, and the experimental results demonstrate its stronger stability and higher success rate on complex tasks.
    \end{itemize}

\section{Related Work}

\subsection{Vision-Based Reinforcement Learning}

    Vision-based policies show greater potential for real-world applications because they can prevent the policy from becoming overly reliant on privileged states, but they are more difficult to train due to partial observability and the high dimensionality of image inputs.
    To date, there have been numerous studies on how to train vision-based policies. 
    Asymmetric actor critic \cite{Pinto_2017_Asymmetric_actor} leverages state-based critics to provide target values for vision-based actors, avoiding difficult vision-based critic learning. Nevertheless, asymmetric training may lead to an unbalanced model learning trajectory.
    To ensure that both the actor and the critic receive image inputs, Yarats et al. \cite{Kostrikov_2020_Image_augmentation, Yarats_2021_Mastering_visual} developed DrQ and DrQv2 for image-based RL training by introducing an augmentation technique. 
    Laskin et al. \cite{Laskin_2020_Curl_Contrastive} proposed a method based on contrastive representation learning.
    With sufficient expert demonstrations, imitation learning can obtain policies via supervised learning \cite{Fu_2024_Mobile_aloha, Chi_2025_Diffusion_policy}, 
    whereas the high demand for expert demonstrations is a limitation in the context of surgical robotic manipulation, where expert demonstrations are more expensive and scarce.
    These methods all achieve great performance in humanoid robotic control, but are challenged when applied to surgical robotic manipulation.

\subsection{Learning-Based Surgical Robotic Manipulation}

    To date, most methods focus on state-based policies, where the robotic agent is assumed to have access to the privileged states of objects \cite{Bendikas_2023_Learning_needle, Azad_2026_SurgiDiff_A, Long_2023_Human_in, Wang_2026_DIRA_Diffusion, Huang_2023_Guided_Reinforcement, Huang_2023_Value_informed}.
    Based on state-based policies, surgical robotic manipulation is inevitably solved in two steps: perception and then manipulation. 
    For example, the agent acquires object states via a hand-crafted image processing pipeline and then feeds the states into the policy network to obtain the robotic actions \cite{Wilcox_2022_Learning_to}.
    Such working pipelines are intuitive and simple to deploy in the real world.
    Nevertheless, a pure vision-based policy is still an attractive solution because of its one-step processing and stronger robustness against noise and varying scenes. 
    To this end, some studies start to involve pixel-based observations \cite{Haiderbhai_2024_Sim2real_rope, Duan_2026_Real_Time, Gondokaryono_2023_Learning_nonprehensile, He_2025_Cosmos_H}.  
    Haiderbhai et al. \cite{Haiderbhai_2024_Sim2real_rope} and Gondokaryono et al. \cite{Gondokaryono_2023_Learning_nonprehensile} trained RGB-based policies with conventional CNNs and LSTMs for basic manipulation tasks.
    To deal with more complex tasks, most vision-based studies depend on additional input observations, such as depth and segmentation masks, which may require extra processing pipelines, or rely on pretrained large models, resulting in high computational costs and potential error propagation.
    In this case, how to efficiently obtain a simple and effective vision-based policy remains a problem.

\subsection{Policy Distillation}

    The concept of policy distillation was first introduced by Rusu et al. \cite{Rusu_2015_Policy_distillation}, aiming to transfer the knowledge of multiple task-specific deep Q-networks to a single network. 
    Czarnecki et al. \cite{Czarnecki_2019_Distilling_Policy} further pointed out that the learning trajectory can oscillate once reward learning is involved in the distillation process.
    Based on policy distillation, various vision-based policies for real-world manipulation tasks have been developed, by using Kullback-Leibler (KL) divergence or Mean Square Error (MSE) to transfer policy knowledge \cite{Levine_2016_End_to, Liu_2022_Distilling_Motion, Chen_2024_Trakdis_A}.
    In most settings, policy distillation employs one policy as the teacher and another as the student, which can largely avoid the high demand for expert demonstrations, and trains the student policy through relatively stable supervised learning.
    We identify that the above properties of policy distillation are more appropriate for surgical vision-based policy training (i.e., low demand for expert demonstrations and a comparatively smooth training trajectory). Furthermore, most studies pertain to discrete action spaces, which cannot be applied to continuous-control laparoscopic robotic manipulation.
    These considerations motivate our study.

    \begin{figure*}[t]
    \centering
    \includegraphics[width=.8\linewidth]{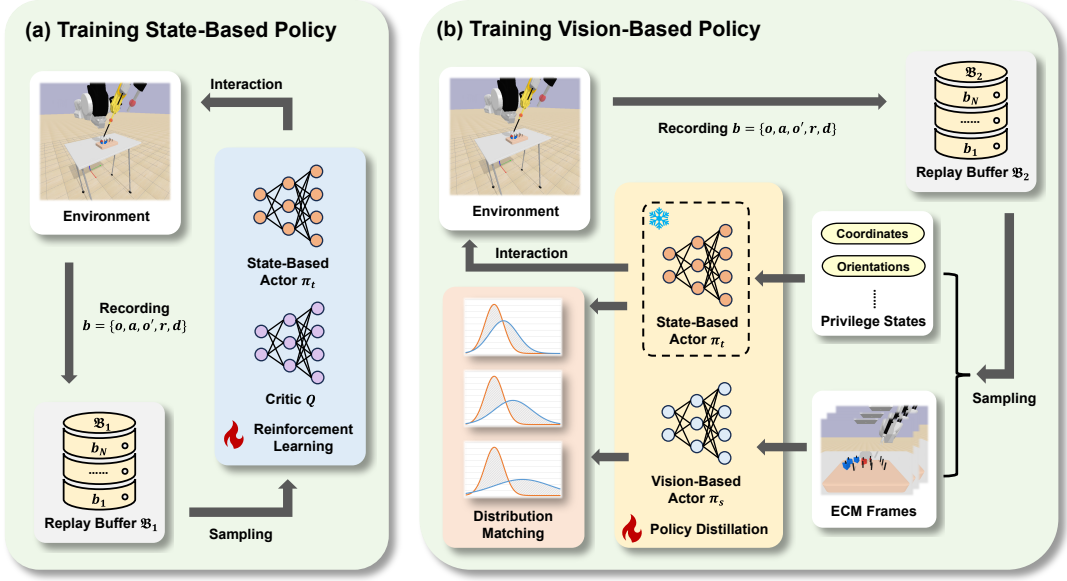}
    \caption{Semantic decomposition of this work. (a) The first step is to train a state-based policy. (b) Then, the state-based policy is employed as the teacher to transfer policy knowledge to the student network by aligning their action distributions. A new replay buffer $\mathfrak{B}_2$ is used in (b) because $\mathfrak{B}_1$ in (a) contains transitions collected by earlier versions of the teacher policy, which may interfere with the student policy's learning of the final teacher policy.}
    \label{fig_method}
    \end{figure*}

\section{Methods}

    The overall workflow of this study is displayed in Fig.~\ref{fig_method}. 
    Specifically, when training a vision-based policy (see Fig.~\ref{fig_method}(b)), we employ a frozen teacher policy to interact with the environment (i.e., teacher-driven distillation) for its relatively stronger stability and reproducibility \cite{Ave_2024_Policy_compression}.

    \subsection{Background}

        Soft Actor-Critic (SAC) \cite{Haarnoja_2018_Soft_actor} is an off-policy RL algorithm that learns a stochastic policy to maximize the expected cumulative return. It learns a policy $\pi_\psi$ and critics $Q_{\phi_1}$ and $Q_{\phi_2}$, where $\psi$ and $\phi_i$ are trainable network parameters. The parameter $\phi_i$ is optimized by minimizing the Bellman error:
        \begin{equation}
            \label{equ_bg_bellman}
            \mathcal{L}(\phi_i, \mathfrak{B}) = \mathbb{E}_{b \sim \mathfrak{B}} \left[ \left( Q_{\phi_i} (o, a) - ( r + \gamma (1 - d) \mathcal{T} ) \right)^2 \right],
        \end{equation}
        where $b = (o, a, o', r, d)$ is a transition with observation $o$, action $a$, next observation $o'$, reward $r$ and done signal $d$, $\mathfrak{B}$ represents the replay buffer, $\gamma$ is the discount factor, and $\mathcal{T}$ is the target value, defined as:

        \begin{equation}
            \label{equ_bg_target_value}
            \mathcal{T} = \min_{i=1,2} Q^*_{\phi_i} (o', \tilde{a}') - \alpha \log \pi_\psi (\tilde{a}' | o'),
        \end{equation}
        where $\tilde{a}'$ indicates the predicted next action sampled from the policy $\pi_\psi (\tilde{a}' | o')$, $Q^*_{\phi_i}$ denotes the exponential moving average (EMA) of the network parameters (i.e., $\phi_i$) of $Q_{\phi_i}$, EMA is empirically employed to improve training stability in off-policy RL algorithms, and the entropy coefficient $\alpha > 0$ determines the relative weighting between entropy maximization and value function optimization.

        In continuous-control tasks, the predicted policy $\pi_\psi$ is typically a Gaussian distribution:
        \begin{equation}
            \label{equ_bg_actor_policy}
            \pi_\psi (\tilde{a} | o) = \mathcal{N}(\mu_\psi(o), \text{diag}(\sigma_\psi^2(o))),
        \end{equation}
        where $\tilde{a}$ is the predicted action given the observation $o$, $\tilde{a}$, $\mu_\psi$ and $\sigma_\psi$ are treated as vectors whose dimensionality corresponds to the number of joints, as the joints are assumed to be conditionally independent of each other.
        With the critic given by $Q_{\phi_i}$, the actor network is trained by maximizing the expected return of its actions as follows:
        \begin{equation}
            \label{equ_bg_actor_loss}
            \mathcal{L}(\psi) = \mathbb{E}_{a \sim \pi} \left[ \min_{i=1,2} Q_{\phi_i} (o, \tilde{a}) - \alpha \log \pi (\tilde{a} | o) \right],
        \end{equation}
        where action $\tilde{a}$ is sampled stochastically from the policy via reparameterization for gradient propagation:
        \begin{equation}
            \label{equ_bg_actor_random_sample}
            \tilde{a} (o, \xi) = \tanh \left(\mu_\psi(o) + \sigma_\psi(o) \odot \xi \right),
        \end{equation}
        where $\xi \sim \mathcal{N}(0, I)$ is a standard Gaussian noise vector. In continuous-control reinforcement learning, stochastic policies are commonly parameterized as factorized Gaussian distributions, where the action dimensions are assumed to be conditionally independent. This parameterization provides tractable computation of policy likelihoods and facilitates efficient reparameterized sampling.

    \subsection{State-to-Vision Policy Distillation}
        We revisit the SAC algorithm and observe that its learned Gaussian policy is particularly well-suited to distillation. Unlike conventional distillation methods that compute various divergence measures via Monte-Carlo estimators, the SAC policy explicitly models both the mean $\mu$ and the standard deviation $\sigma$, enabling closed-form distillation between teacher and student policies.
        Therefore, the discrepancy between the teacher and student policies can be accurately measured in the distribution space, providing a more deterministic and interpretable optimization.
        Moreover, the predicted standard deviation explicitly characterizes the uncertainty of the policy, allowing the student to learn both the action tendency and the confidence of the teacher behavior. Such uncertainty-aware policy transfer may lead to learning behaviors that differ from those of the conventional mean-only transfer.

        For convenience and readability, the subsequent derivations are presented in terms of one joint, because different joints are assumed to be conditionally independent. First, we define the teacher policy and the student policy as:
        \begin{equation}
            \label{equ_s2v_st_pi}
            \begin{split}
                \pi_t (\tilde{a} | o_t) &= \mathcal{N}(\mu_t(o_t), \sigma_t^2(o_t)), 
                \\
                \pi_s (\tilde{a} | o_s) &= \mathcal{N}(\mu_s(o_s), \sigma_s^2(o_s)),
            \end{split}
        \end{equation}
        where the network parameter $\psi$ is replaced with $t$ and $s$ to indicate the state-based network (i.e., teacher) and the RGB-based network (i.e., student), respectively, $o_t$ indicates the privileged state-based observation, and $o_s$ indicates the visual observation (i.e., RGB frames).
        Under challenging laparoscopic robotic manipulation tasks, directly training $\pi_s$ using RL is difficult. Therefore, identifying an effective distillation objective to align the two policies introduces an alternative solution for obtaining an RGB-based policy. The distillation loss is defined as:
        \begin{equation}
            \label{equ_s2v_dist_loss}
            \mathcal{L}_\text{distill} = D(\pi_t, \pi_s),
        \end{equation}
        where $D$ is a differentiable function that measures the difference between $\pi_t$ and $\pi_s$. Given a pre-trained and frozen state-based policy $\pi_t$, optimizing the visual policy $\pi_s$ by minimizing Eq.~\eqref{equ_s2v_dist_loss} can transfer policy knowledge from $\pi_t$ to $\pi_s$.

    \subsection{Distribution Matching Objectives}
        Given the stochastic Gaussian policies of the teacher and student, state-to-vision policy transfer can be formulated as a distribution matching problem. Specifically, we seek to minimize a divergence or distance between the two policies, such as the Wasserstein Distance (WD) and the Kullback-Leibler (KL) Divergence.
        Forward KL (FKL) divergence is a prevalent technique in various deep learning fields, such as model distillation, knowledge transfer, and representation alignment. Its Gaussian closed form is defined as:
        \begin{equation}
            \label{equ_fkl_eq}
            \begin{split}
            D_\text{FKL}(\pi_t, \pi_s) &= D_\text{KL}(\pi_t || \pi_s) \\
            &= \log(\frac{\sigma_s}{\sigma_t}) + \frac{\sigma_t^2 + (\mu_t - \mu_s)^2}{2 \sigma_s^2} - 0.5.
            \end{split}
        \end{equation}
        Similarly, the Reverse KL (RKL) divergence between the state-based policy and the vision-based policy is written as:
        \begin{equation}
            \label{equ_rkl_eq}
            D_\text{RKL}(\pi_t, \pi_s) = D_\text{KL}(\pi_s || \pi_t).
        \end{equation}
        Although RKL is not a prevalently used method for distillation, we investigate it to gain more insights into how the uncertainty influences the learning behavior.
        FKL and RKL measure the discrepancy between the teacher and student policies, but their asymmetric formulations lead to distinct optimization behaviors. FKL exhibits a mode-covering tendency, which may encourage the student policy to increase its variance to cover the probability mass of the teacher distribution. In contrast, RKL exhibits a mode-seeking tendency, favoring concentrated distributions and potentially underestimating the uncertainty of the policy action \cite{Czarnecki_2019_Distilling_Policy, GX_Chen_2026_KL_regularized}. To balance these complementary behaviors, we further consider the Jeffreys divergence (JD), also known as the symmetric KL divergence (SKL), which incorporates both forward and reverse KL divergences:
        \begin{equation}
            \label{equ_jd_1}
            \begin{split}
                D_\text{JD}(\pi_t, \pi_s) &= D_{KL} (\pi_t || \pi_s) + D_{KL} (\pi_s || \pi_t) \\
                &= \frac{1}{2} \cdot \Big( \frac{\sigma_s^2}{\sigma_t^2} + \frac{\sigma_t^2}{\sigma_s^2} + \frac{\left( \mu_t - \mu_s \right)^2}{\sigma_t^2} \\
                & \quad + \frac{\left( \mu_t - \mu_s \right)^2}{\sigma_s^2} \Big) - 1.
            \end{split}
        \end{equation}
        

        In addition to the aforementioned information-theoretic divergences, we also examine the Wasserstein Distance (WD), a geometrically grounded discrepancy based on optimal transport. For two univariate Gaussian distributions, the squared 2-Wasserstein distance admits the following closed-form expression \cite{Gelbrich_1990_On_a}:
        \begin{equation}
            \label{equ_wd_eq}
            D_\text{WD}(\pi_t, \pi_s) = W_2^2(\pi_t, \pi_s) = (\mu_t - \mu_s)^2 + (\sigma_t - \sigma_s)^2.
        \end{equation}
        This formulation explicitly accounts for discrepancies in both the mean and the standard deviation of the two policy distributions. Interestingly, the mean-matching term in Eq.~\ref{equ_wd_eq} is identical to that of the conventional Mean-Square Error (MSE) distillation loss:
        \begin{equation}
            \label{equ_mse_eq}
            D_\text{MSE}(\pi_t, \pi_s) = (\mu_t - \mu_s)^2,
        \end{equation}
        which considers only the discrepancy between the means. 
        
        
        Different distillation losses exhibit distinct optimization behaviors and advantages. For instance, WD, MSE, and RKL exhibit strictly convex optimization landscapes with respect to $\mu_s$ and $\sigma_s$, while FKL and JD introduce uncertainty-adaptive optimization through standard-deviation scaling of the gradients, which is analyzed in Section \ref{Sec_grad_analysis}.

    \subsection{Gradient Analysis}
    \label{Sec_grad_analysis}

        \begin{table}[t]
            \centering
            \caption{First- and second-order derivatives of different distribution matching objectives with respect to the visual policy parameters $\mu_s$ and $\sigma_s$. We define $\Delta\mu=\mu_s-\mu_t$. The row for $\frac{\partial D}{\partial \sigma_s \mu_s}$ is omitted because $\frac{\partial^2 D}{\partial \sigma_s \mu_s} = \frac{\partial^2 D}{\partial \mu_s \sigma_s}$.}
            \label{tab_loss_derivatives}
            \renewcommand{\arraystretch}{2.0}
            \resizebox{.9\linewidth}{!}{
            \begin{tabular}{c c c c c c}
                \toprule
                \textbf{Derivative}
                & \textbf{FKL}
                & \textbf{RKL}
                & \textbf{JD}
                & \textbf{WD}
                & \textbf{MSE} \\
                \midrule

                $\displaystyle\frac{\partial D}{\partial \mu_s}$
                &
                $\displaystyle
                \frac{\Delta\mu}{\sigma_s^2}$
                &
                $\displaystyle
                \frac{\Delta\mu}{\sigma_t^2}$
                &
                $\displaystyle
                \Delta\mu
                \left(
                \frac{1}{\sigma_s^2}
                +\frac{1}{\sigma_t^2}
                \right)$
                &
                $\displaystyle
                2\Delta\mu$
                &
                $\displaystyle
                2\Delta\mu$
                \\

                $\displaystyle\frac{\partial D}{\partial \sigma_s}$
                &
                $\displaystyle
                \frac{1}{\sigma_s}
                -\frac{\sigma_t^2+\Delta\mu^2}{\sigma_s^3}$
                &
                $\displaystyle
                -\frac{1}{\sigma_s}
                +\frac{\sigma_s}{\sigma_t^2}$
                &
                $\displaystyle
                \frac{\sigma_s}{\sigma_t^2}
                -\frac{\sigma_t^2+\Delta\mu^2}
                {\sigma_s^3}$
                &
                $\displaystyle
                2(\sigma_s-\sigma_t)$
                &
                $0$
                \\

                $\displaystyle\frac{\partial^2 D}{\partial \mu_s^2}$
                &
                $\displaystyle
                \frac{1}{\sigma_s^2}$
                &
                $\displaystyle
                \frac{1}{\sigma_t^2}$
                &
                $\displaystyle
                \frac{1}{\sigma_s^2}
                +\frac{1}{\sigma_t^2}$
                &
                $2$
                &
                $2$
                \\

                $\displaystyle\frac{\partial^2 D}{\partial \sigma_s^2}$
                &
                $\displaystyle
                -\frac{1}{\sigma_s^2}
                +\frac{3(\sigma_t^2+\Delta\mu^2)}
                {\sigma_s^4}$
                &
                $\displaystyle
                \frac{1}{\sigma_s^2}
                +\frac{1}{\sigma_t^2}$
                &
                $\displaystyle
                \frac{1}{\sigma_t^2}
                +\frac{3(\sigma_t^2+\Delta\mu^2)}
                {\sigma_s^4}$
                &
                $2$
                &
                $0$
                \\

                $\displaystyle\frac{\partial^2 D}
                {\partial \mu_s\partial\sigma_s}$
                &
                $\displaystyle
                -\frac{2\Delta\mu}{\sigma_s^3}$
                &
                $0$
                &
                $\displaystyle
                -\frac{2\Delta\mu}{\sigma_s^3}$
                &
                $0$
                &
                $0$
                \\

                \bottomrule
            \end{tabular}
            }
        \end{table}

        The first- and second-order derivatives of the different distribution matching objectives are summarized in Table~\ref{tab_loss_derivatives}. 
        As shown in the first row for $\frac{\partial D}{\partial \mu_s}$, all objectives drive the visual policy mean $\mu_s$ toward the state policy mean $\mu_t$. Furthermore, for FKL, RKL, and JD, the magnitude of the resulting gradient is additionally modulated by the standard deviations of the two policies, leading to different degrees of adaptive scaling.

        
        In particular, FKL, RKL, and JD assign larger gradient magnitudes to policy dimensions with lower uncertainty, i.e., smaller standard deviations. This property enables the optimization of $\mu_s$ to adapt to the confidence of the corresponding policy distributions.
        In contrast, the gradients of $\mu_s$ for both WD and MSE are independent of $\sigma_t$ and $\sigma_s$. 
        Consequently, their optimization of the policy mean follows the same objective, as both minimize the squared difference between $\mu_s$ and $\mu_t$. 
        Although WD additionally minimizes the discrepancy between $\sigma_s$ and $\sigma_t$, under teacher-driven distillation, this variance-matching term has minimal effect on the gradient of $\mu_s$ or the interaction with the environment. 
        Therefore, for teacher-driven distillation, with respect to the optimization of $\mu_s$, WD is similar to MSE.


    
\begin{figure}[t]
\centering
\includegraphics[width=.8\linewidth]{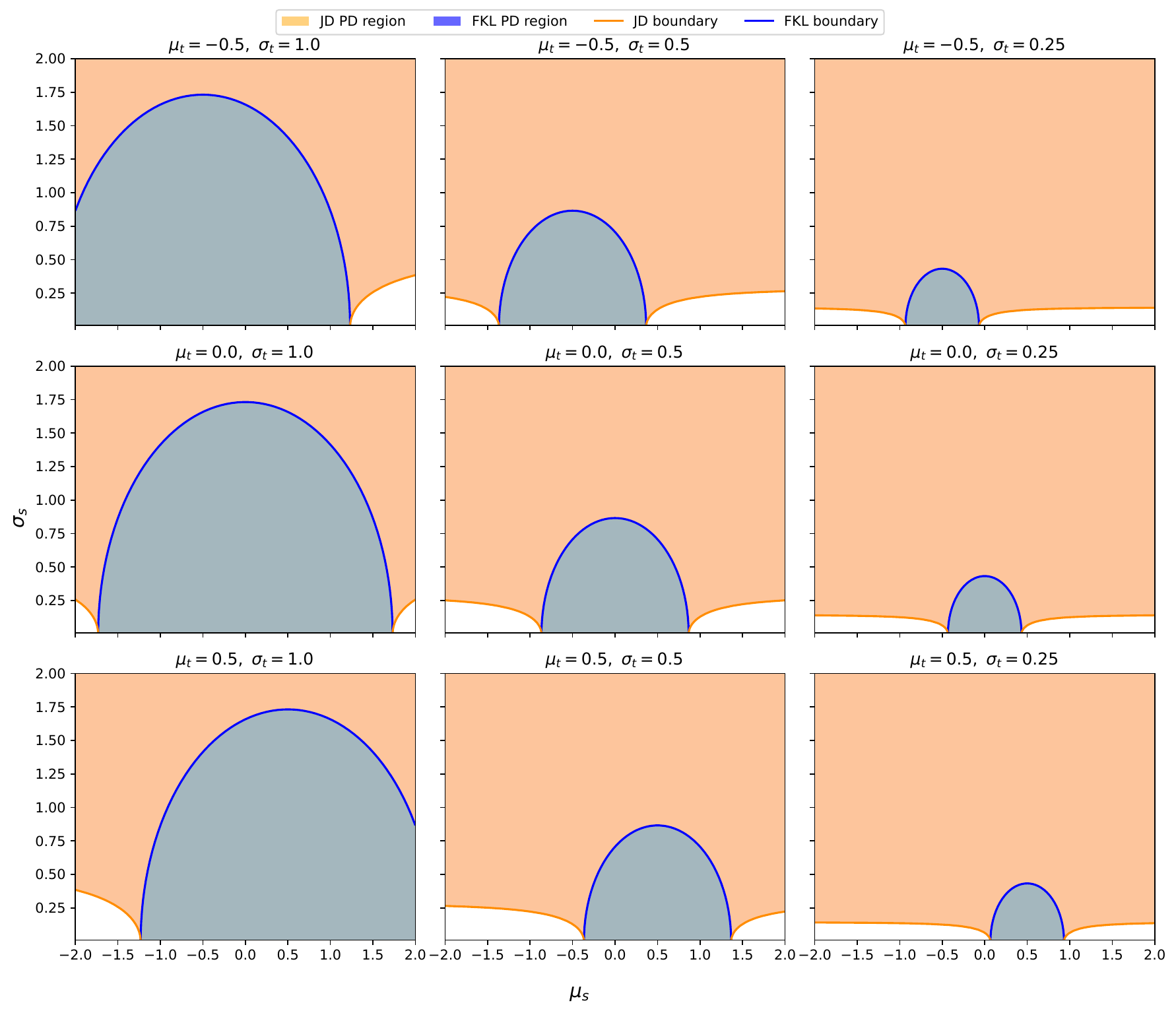}
\caption{Toy illustration of the PD regions of the FKL and JD Hessians under different teacher policy parameters. The PD region of FKL is always a proper subset of that of JD.}
\label{fig_fkl_jd_pd_33}
\end{figure}

\begin{table*}[t]
    \centering
    \caption{Success Rate (\%) of Various Methods Across Different Tasks. The "Learning" column indicates Reinforcement Learning (RL), Imitation Learning (IL), or Policy Distillation (PDL). \textbf{Bold} marks the best result for each task across different methods. \underline{Underline} denotes the second-best one. Results marked with "*" are directly copied from original publications.}
    \label{tab_perf_comp}
    \resizebox{.8\textwidth}{!}{
    \begin{tabular}{l c c c c c c}
        \toprule
        Method & Learning & NeedleReach & GauzeRetrieve & NeedlePick & PegTransfer & NeedleRegrasp \\ 
        \midrule
        CURL \cite{Laskin_2020_Curl_Contrastive} & RL & \underline{98.8} & 1.8 & 0.0 & 0.0 & 0.0 \\ 
        DrQv2 \cite{Yarats_2021_Mastering_visual} & RL & \textbf{99.8} & 2.8 & 1.2 & 0.0 & 0.0 \\ 
        Diffusion \cite{Chi_2025_Diffusion_policy} & RL & 80.2 & 13.4 & 22.4 & 2.2 & 1.4 \\ 
        ACT \cite{Duan_2026_Real_Time} & IL & N/A & N/A & $76.0^*$ & N/A & N/A \\
        ALOHA \cite{Fu_2024_Mobile_aloha} & IL & \textbf{99.8} & \textbf{76.8} & \textbf{93.6} & 16.4 & 10.0 \\ 
        \midrule
        Ours (FKL) & PDL & 98.6 & 66.0 & 70.8 & \underline{75.4} & \underline{71.6} \\ 
        Ours (RKL) & PDL & \textbf{99.8} & 76.8 & 39.8 & 74.8 & 70.2 \\ 
        Ours (JD) & PDL & \underline{98.8} & \underline{71.8} & \underline{80.4} & \textbf{77.4} & \textbf{74.8} \\ 
        Ours (MSE) & PDL & 96.8 & 69.8 & 74.2 & 73.8 & 69.8 \\
        \bottomrule
    \end{tabular}
    }
\end{table*}

    \subsection{Positive-Definiteness of Different Hessian Matrices}
    \label{sec_pd_hm}

        The Hessian matrix provides a useful perspective for understanding the optimization behavior of different distribution matching objectives. 
        A positive-definite (PD) Hessian indicates that the objective exhibits fully positive curvature in all directions, corresponding to a locally strictly convex optimization landscape around the considered point.
        The Hessian matrix with respect to $\mu_s$ and $\sigma_s$ is defined as:
        \begin{equation}
            H_D
            =
            \nabla^2 D
            =
            \begin{bmatrix}
                \displaystyle
                \frac{\partial^2 D}{\partial u_s^2}
                &
                \displaystyle
                \frac{\partial^2 D}{\partial u_s \partial \sigma_s}
                \\[8pt]
                \displaystyle
                \frac{\partial^2 D}{\partial \sigma_s \partial u_s}
                &
                \displaystyle
                \frac{\partial^2 D}{\partial v_s^2}
            \end{bmatrix}.
        \end{equation}

        The RKL, WD and MSE Hessian matrices are clearly positive definite, whereas parameters $\mu_s$ and $\sigma_s$ are decoupled, i.e., $\partial^2 D_\text{RKL} / (\partial \mu_s \partial \sigma_s) = 0$, which means that updating $\sigma_s$ does not affect the optimization of $\mu_s$.
        Consequently, when training the student policy, RKL is similar to a prior-scaled WD or MSE objective, where the gradient with respect to $\mu_s$ is scaled by a fixed factor $\sigma_t^{-2}$.

        For FKL and JD, the Hessian matrices are not always positive definite because their determinants are not greater than zero for all parameters:
        \begin{equation}
        \label{Equ_FKL_Det}
            \text{det}(H_\text{FKL}) = \sigma_s^{-6} \cdot \left( 3 \sigma_t^2 - (\Delta \mu)^2 - \sigma_s^2 \right),
        \end{equation}
        \begin{equation}
        \label{Equ_JD_Det}
          \begin{split}
            \text{det}(H_\text{JD}) &= \left( \frac{1}{\sigma_t^2} + \frac{1}{\sigma_s^2} \right) \cdot \left( \frac{1}{\sigma_t^2} + \frac{3 \left( \sigma_t^2 + (\Delta \mu)^2 \ \right)}{\sigma_s^4} \right) \\
            & \quad - \frac{4 (\Delta \mu)^2}{\sigma_s^6}.
          \end{split}
        \end{equation}
        This indicates that their Hessian matrices can be indefinite and that the optimization landscape is non-convex in certain directions. 
        On the other hand, FKL and JD introduce adaptively scaled gradients with respect to $\mu_s$, where the scaling factor is determined by the student uncertainty in FKL and jointly by the teacher and student uncertainties in JD, resulting in uncertainty-aware optimization behavior. 
        In contrast, RKL scales the gradient with respect to $\mu_s$ solely by the teacher uncertainty, which remains fixed during student optimization.


        Although neither the FKL nor the JD Hessian is positive definite over its entire domain, the JD Hessian exhibits a broader domain of positive definiteness than the FKL Hessian. We analyze this in Section \ref{Sec_relaxed_PD_JD}.

        \subsection{Relaxed Positive-Definiteness Condition}
        \label{Sec_relaxed_PD_JD}

        Let $\boldsymbol{x}=[\mu_t, \sigma_t, \mu_s, \sigma_s]^T$ denote parameters of two policies, and let:
        \begin{equation}
        H_{\mathrm{FKL}}(\boldsymbol{x})
        =
        \nabla^2 D_{\mathrm{FKL}}(\boldsymbol{x}),
        \;
        H_{\mathrm{RKL}}(\boldsymbol{x})
        =
        \nabla^2 D_{\mathrm{RKL}}(\boldsymbol{x})
        \end{equation}
        denote the corresponding Hessian matrices. The JD is given by Eq.~\eqref{equ_jd_1} and therefore its Hessian satisfies:
        \begin{equation}
        H_{\mathrm{JD}}
        =
        H_{\mathrm{FKL}}+
        H_{\mathrm{RKL}}.
        \label{eq:jd_hessian_decomposition}
        \end{equation}


        We have previously established that the RKL Hessian is strictly positive definite, i.e.,
        \begin{equation}
        H_{\mathrm{RKL}}(\boldsymbol{x}) \succ 0,
        \end{equation}
        throughout the considered parameter domain. Consequently, for any nonzero vector $\boldsymbol{z}$:
        \begin{equation}
        \boldsymbol{z}^{\top}
        H_{\mathrm{RKL}}(\boldsymbol{x})
        \boldsymbol{z}
        >0.
        \label{eq:rkl_pd}
        \end{equation}
        This property leads directly to the following proposition.


        \textbf{Proposition 1.}
        For any $\boldsymbol{x}$ satisfying $H_{\text{FKL}}(\boldsymbol{x}) \succ 0$, we have $H_{\text{JD}}(\boldsymbol{x}) \succ 0$. Moreover, there exists an $\boldsymbol{x}_0$ satisfying $H_{\text{JD}}(\boldsymbol{x}_0) \succ 0$ while $H_{\text{FKL}}(\boldsymbol{x}_0) \nsucc 0$. Therefore, the PD region of FKL is a proper subset of that of JD:
        \begin{equation}
        \boxed{
        \mathcal{P}_{\mathrm{FKL}}
        \varsubsetneqq 
        \mathcal{P}_{\mathrm{JD}},
        }
        \label{eq:pd_region_inclusion}
        \end{equation}
        \begin{equation}
        \mathcal{P}_{\mathrm{FKL}}
        =
        {
        \boldsymbol{x}:
        H_{\mathrm{FKL}}(\boldsymbol{x})\succ0
        },
        \;
        \mathcal{P}_{\mathrm{JD}}
        =
        {
        \boldsymbol{x}:
        H_{\mathrm{JD}}(\boldsymbol{x})\succ0
        }.
        \end{equation}
        
        \begin{proof}
        Consider an arbitrary $\boldsymbol{x} \in \mathcal{P}_{\mathrm{FKL}}$. By definition, we have $H_{\mathrm{FKL}}(\boldsymbol{x})\succ0$.
        For any nonzero vector $\boldsymbol{z}$, we therefore obtain:
        \begin{equation}
        \label{equ_fkl_pd}
        \boldsymbol{z}^{\top}
        H_{\mathrm{FKL}}(\boldsymbol{x})
        \boldsymbol{z}
        >0.
        \end{equation}
        Using Eqs.~\eqref{eq:jd_hessian_decomposition}, \eqref{eq:rkl_pd}, and \eqref{equ_fkl_pd}, we have:
        \begin{equation}
            \begin{split}
              \boldsymbol{z}^{\top}
              H_{\mathrm{JD}}(\boldsymbol{x})
              \boldsymbol{z}
              &=
              \boldsymbol{z}^{\top}
              \left(
              H_{\mathrm{FKL}}(\boldsymbol{x})
              +
              H_{\mathrm{RKL}}(\boldsymbol{x})
              \right)
              \boldsymbol{z}
              \\
              &=
              \boldsymbol{z}^{\top}
              H_{\mathrm{FKL}}(\boldsymbol{x})
              \boldsymbol{z}
              +
              \boldsymbol{z}^{\top}
              H_{\mathrm{RKL}}(\boldsymbol{x})
              \boldsymbol{z}
              >0.
            \end{split}
        \end{equation}
        Hence, $H_{\mathrm{JD}}(\boldsymbol{x})\succ0$.
        Since $\boldsymbol{x}$ is chosen arbitrarily from $\mathcal{P}_{\mathrm{FKL}}$, we obtain
        $\mathcal{P}_{\mathrm{FKL}}
        \subseteq
        \mathcal{P}_{\mathrm{JD}}$.
        Specifically, by setting $\mu_t = 0$ and $\sigma_t = 0.5$, we consider the student parameters $\mu_s = 0.75$ and $\sigma_s = 1.0$, i.e., existing a point $\boldsymbol{x}_0 = [0, 0.5, 0.75, 1.0]^T$. 
        At this point, using Eqs.~\eqref{Equ_FKL_Det} and \eqref{Equ_JD_Det}, we have $\text{det}(H_\text{FKL}(\boldsymbol{x}_0)) = -0.8125 < 0 \implies H_\text{FKL}(\boldsymbol{x}_0) \nsucc 0$, and $\text{det}(H_\text{JD}(\boldsymbol{x}_0)) = 29.9375 > 0 \implies H_\text{JD}(\boldsymbol{x}_0) \succ 0$. 
        Therefore, we have:
        \begin{equation}
        \mathcal{P}_{\mathrm{FKL}} \subsetneqq \mathcal{P}_{\mathrm{JD}}.
        \end{equation}
        This proves the proposition. 
        \end{proof}

        The above result establishes that the PD region of FKL is a proper subset of that of JD, indicating that JD admits a larger region of local convexity. We provide a toy illustration in Fig.~\ref{fig_fkl_jd_pd_33}. 
        Therefore, JD does not merely symmetrize the divergence and balance mode covering and mode seeking. It also introduces a more relaxed positive-definiteness condition than FKL and can exhibit more favorable local optimization.

\section{Experiments and Results}
\label{Sec_exp_res}

    \subsection{Implementation Details}
    \label{Sec_imp_det}

    All experiments are conducted in the SurRoL simulator \cite{Xu_2021_Surrol_An}.
    The teacher policy receives privileged state information, including the coordinates and orientations of objects such as the PSM, needles, and gauze, whereas the student policy takes RGB frames from the ECM along with only the PSM gripper coordinates, without access to the states of other objects.
    All state policies are implemented as three-layer MLPs, and all RGB policies are implemented with three convolutional layers and three fully connected layers similar to those of CURL \cite{Laskin_2020_Curl_Contrastive} and DrQ \cite{Kostrikov_2020_Image_augmentation}. 
    The teacher policy is first trained using Soft Actor-Critic (SAC) with the default hyperparameter settings \cite{Haarnoja_2018_Soft_actor}, including the learning rate, the discount factor, etc.
    After training, the teacher policy is frozen and used
    to provide supervision for training the vision-based student policy.
    In particular, the JD loss is scaled by $0.5$ to align its gradient magnitude with those of the other distribution matching objectives for fairness. 
    To reduce experimental time costs, WD is not implemented separately because it shares the similar learning objective as MSE. 
    The success rate for each evaluation is reported over 100 independent trials. 
    Each experiment is independently repeated five times using different random seeds. 
    All results are reported as averages across the five runs.
    Student policies trained by different loss distillation methods utilize the same teacher policy.
    All methods are trained via the Adam optimizer with a learning rate of $10^{-4}$, a batch size of 64, and $5 \times 10^{5}$ iterations.
    During training, we periodically evaluate the policy every 1000 iterations and save the checkpoint with the highest reward for subsequent evaluation.
    For more implementation details, please visit \href{https://github.com/MingyuShengSMY/SurgiPD.git}{https://github.com/MingyuShengSMY/SurgiPD.git}.

\begin{figure*}[t]
\centering
\includegraphics[width=0.8\linewidth]{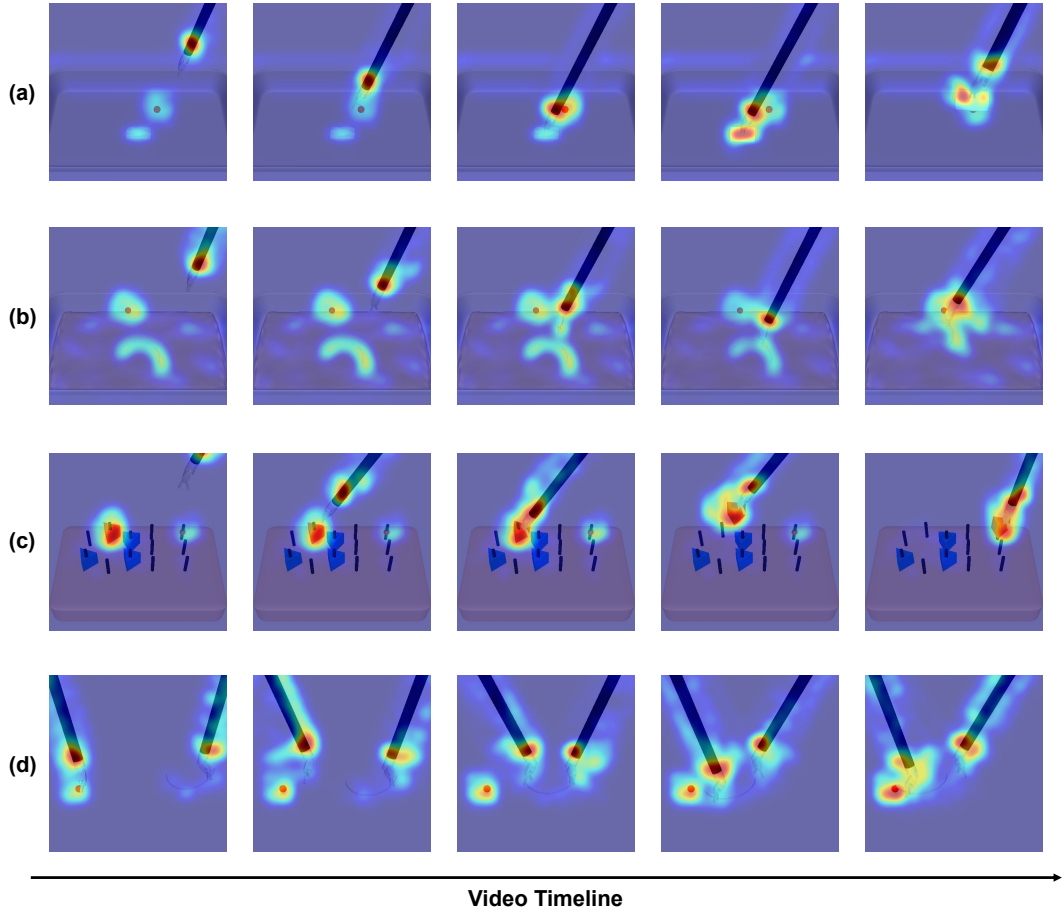}
\caption{Feature maps extracted from the last convolutional layer. (a), (b), (c), and (d) represent GauzeRetrieve, NeedlePick, PegTransfer, and NeedleRegrasp, respectively. The small red orb marks the position to which the target objects (e.g., needle and gauze) should be moved. For a human-friendly view, the raw feature maps ($64 \times 64 \times 32$) are averaged across the channel dimension and refined via Gaussian blurring and interpolation.}
\label{fig_feat_map}
\end{figure*}

    \subsection{Results Comparison Across Different Tasks}
    \label{Sec_exp_res_sota_comp}

    Success rates of various methods under different tasks are shown in Table~\ref{tab_perf_comp}.  
    For pure RL-based vision methods (e.g., CURL and DrQv2), their performances are good on the simplest task (i.e., NeedleReach), whereas their effectiveness is significantly challenged when facing more complex tasks, such as NeedlePick and NeedleRegrasp where more dexterous manipulations are involved.
    ALOHA demonstrates superior performance on NeedleReach, GauzeRetrieve, and NeedlePick, but also exhibits performance degradation on more challenging tasks (i.e., PegTransfer and NeedleRegrasp).
    In terms of policy distillation, although its performance is similar to that of other methods on the simple manipulation task (i.e., NeedleReach), and slightly worse than ALOHA on GauzeRetrieve and NeedlePick, it can maintain a success rate of about 70\% on complex manipulation tasks, showing relatively stronger stability and robustness. 


    \subsection{Results Comparison Across Different Distillations}
    \label{sec_res_comp_dist}

    To investigate the contributions of different distillation settings, we evaluate and compare the performances of FKL, RKL, JD, and MSE across various tasks (see Table~\ref{tab_perf_comp}).
    Specifically, noticeably inferior performance is observed for RKL on NeedlePick, at 39.8\%, showing its potential instability. 
    Recalling the gradient equation of RKL in Table~\ref{tab_loss_derivatives}, we hypothesize that the instability may result from the combined effects of the prior-scaled gradient (discussed in Section~\ref{sec_pd_hm}) and the high certainty of the teacher policy, whose standard deviation is empirically around 0.3 in our experiments.
    In this case, the gradient is amplified threefold, and thus may cause gradient oscillation and lower performance.


    Uncertainty-adaptive distillations (i.e., FKL and JD) usually outperform non-adaptive ones (i.e., RKL and MSE) on complex tasks (see PegTransfer and NeedleRegrasp in Table~\ref{tab_perf_comp}).
    On the other hand, for simpler tasks, such as NeedleReach and GauzeRetrieve, there is no significant enhancement gained from uncertainty adaptation. 
    This observation suggests the potential effectiveness of uncertainty adaptation for complex laparoscopic manipulation tasks.


    Comparing FKL and JD, the performance of JD is slightly better than that of FKL across different tasks to varying degrees, corresponding to the Hessian matrix analysis in Section \ref{Sec_relaxed_PD_JD}, where JD features a broader convex optimization landscape that benefits the training.

    \subsection{Feature Map Analysis}
    \label{sec_feat_map}

    To understand what visual cues the policy relies on, we visualize the CNN feature maps, i.e., the output of the last convolutional layer. 
    As shown in Fig.~\ref{fig_feat_map}, high activations consistently concentrate on the gripper and the target object, while background regions remain weakly activated. 
    Moreover, these high-activation regions follow the gripper and object as their positions change, rather than staying fixed to image coordinates. 
    This indicates that the CNN learns task-relevant semantic representations encoding the relative spatial configuration between the gripper and the object, explaining for the robustness to object position changes.

    \subsection{Conclusion}
    \label{sec_conclusion}

    In this study, we propose a policy distillation framework for visuomotor laparoscopic robotic manipulation tasks, which can transfer policy knowledge from a state-based policy to a vision-based policy through distribution matching objectives.
    Our experimental results in the SurRoL simulator demonstrate that our method shows superior performance and robustness on relatively complex manipulation tasks.
    We also observe the potential effectiveness of the uncertainty adaptation mechanism by comparing different distillation objectives, and investigate the intermediate feature maps to show their capability of task-relevant semantic understanding.
    Although our experiments are conducted in the SurRoL simulator, existing works indicate that transferring such policies to real hardware is viable via sim-to-real methods, such as domain randomization \cite{Xu_2021_Surrol_An, Chiu_2021_Bimanual_Regrasping, Peng_2018_Sim_to}. 
    Furthermore, in this work, we only utilize a simple network architecture consisting of three convolutional layers and three fully connected layers, without relying on advanced mechanisms such as attention, multi-scales, and mixture-of-experts, and only leverage RGB frames and robot posture states as input, indicating the potential scalability to further enhance performance.

\bibliographystyle{IEEEtran} 
\bibliography{main}

\end{document}